\documentclass{article}

\usepackage{microtype}
\usepackage{graphicx}
\usepackage{subcaption}
\usepackage{booktabs} 
\usepackage{multirow}

\usepackage{hyperref}

\usepackage[preprint]{icml2026}

\usepackage{amsmath}
\usepackage{amssymb}
\usepackage{mathtools}
\usepackage{amsthm}

\usepackage[capitalize,noabbrev]{cleveref}

\theoremstyle{plain}
\newtheorem{theorem}{Theorem}[section]

\theoremstyle{definition}

\newtheorem{assumption}[theorem]{Assumption}
\theoremstyle{remark}
\newtheorem{remark}[theorem]{Remark}

\usepackage[textsize=tiny]{todonotes}

\icmltitlerunning{Ren et al., HKD-ACPA: Heterogeneous Knowledge Distillation Adversarial Cyber-Physical Attacks.}

\begin{document}

\twocolumn[
  \icmltitle{HKD-ACPA: Heterogeneous Knowledge Distillation Adversarial Cyber-Physical Attacks in Power Grids}



  \icmlsetsymbol{equal}{*}

  \begin{icmlauthorlist}
    \icmlauthor{Junhao Ren}{equal,ntu}
    \icmlauthor{Honglin Gao}{equal,ntu}
    \icmlauthor{Shuo Feng}{ntu}
    \icmlauthor{Kun Cao}{tju}
    \icmlauthor{Qiyu Kang}{ustc}
    \icmlauthor{Xiaofeng Zong}{cug}
    \icmlauthor{Shuping He}{ahu}
    \icmlauthor{Gaoxi Xiao}{ntu}
  \end{icmlauthorlist}

  \icmlaffiliation{ntu}{School of Electrical and Electronic Engineering, Nanyang Technological University, Singapore}
  \icmlaffiliation{tju}{College of Electronic and Information Engineering, Shanghai Research Institute for Intelligent Autonomous Systems, Tongji University, Shanghai, China}
  \icmlaffiliation{ustc}{School of Information Science and Technology, University of Science and Technology of China, Hefei, China}
  \icmlaffiliation{cug}{School of Automation, China University of Geosciences, Wuhan, China}
  \icmlaffiliation{ahu}{School of Electrical Engineering and Automation, Anhui University, Hefei, China}

  \icmlcorrespondingauthor{Junhao Ren}{junhao002@e.ntu.edu.sg}
  \icmlcorrespondingauthor{Gaoxi Xiao}{egxxiao@ntu.edu.sg}

  \icmlkeywords{Model Distillation, Adversarial Attacks, Cyber-Physical Systems, Power Grids}

  \vskip 0.3in
]



\printAffiliationsAndNotice{}  

\begin{abstract}
  Learning-based algorithms assist power-grid control centers in localizing transmission-line attacks, yet existing adversarial attacks typically require white-box access or repeated online queries. We study a practical zero-query setting in which an adversary coordinates physical, denial-of-service (DoS), and false data injection (FDI) attacks using only local measurements and historical detector outputs. To bridge the input-view asymmetry between local adversarial observations and DoS-masked control-center measurements, we propose Heterogeneous Knowledge Distillation Adversarial Cyber-Physical Attacks (HKD-ACPA). HKD-ACPA distills historical detector predictions into a heterogeneous graph surrogate ensemble and trains a residual calibration module to correct the view-induced logit discrepancy. Over this calibrated surrogate, we formulate sparse, magnitude-bounded FDI synthesis with a group-balanced objective that jointly induces missed detections and false alarms, and solve the resulting hybrid discrete--continuous problem via bi-level projected gradient-based proximal policy optimization (PGPPO). Experiments on the IEEE 30-bus and 118-bus systems show that HKD-ACPA outperforms all surrogate-based baselines across every sparsity budget and surpasses online-query attacks in eight of ten system--budget settings without querying the detector model.
\end{abstract}

\section{Introduction}
Modern smart grids have evolved into complex cyber-physical systems (CPSs) through the tight integration of physical power infrastructures with communication and monitoring platforms, such as Supervisory Control and Data Acquisition (SCADA) systems and Wide-Area Monitoring Systems (WAMS). Although this cyber-physical coupling improves operational efficiency and system observability, it also enlarges the attack surface and introduces new security vulnerabilities \cite{bhattacharjee2024DeeBBAA,sun2025Secure}.

Malicious physical and cyber manipulations can compromise the structural integrity of the grid, disrupt real-time state estimation, and even trigger cascading failures across interconnected infrastructures \cite{ren2026multiscaleattentionbasedattackdiagnosis,fahim2024Graph}. An intelligent adversary may further combine false data injection (FDI), denial-of-service (DoS), and physical attacks to threaten grid operation. To defend against such stealthy and complex threats, control centers increasingly adopt a learning-based detector model, often built with graph neural networks (GNNs), which exploit spatial correlations and topological constraints in power networks to learn discriminative attack patterns from labeled data \cite{fang2023Explainable,boyaci2022Joint,fahim2024Graph}. 

Despite its strong detection performance, a learning-based detector model introduces new attack surfaces and model-specific vulnerabilities \cite{ruan2024On,yu2025GZOO,tian2025EVADE}. Adversaries can exploit these vulnerabilities through carefully designed evasion attacks, for example by injecting small yet effective FDI perturbations that mislead both traditional bad-data detection and the detector model \cite{tian2025EVADE,yu2025GZOO}. However, most existing attack methods rely on strong assumptions. They typically assume either a white-box setting, where the adversary knows the network topology, full measurements, and even internal model details, or an online-query setting, where the deployed detector model can be repeatedly queried so that gradients can be estimated during attack generation. Such assumptions substantially limit the practical applicability of adversarial attack methods in power grids.

In this work, we propose Heterogeneous Knowledge Distillation Adversarial Cyber-Physical Attacks (HKD-ACPA), a zero-query transfer framework in which DoS and FDI attacks are coordinated to undermine physical attack detection and localization on transmission lines by inducing both missed detections and false alarms. HKD-ACPA constructs a calibrated surrogate of the deployed detector model from historical outputs under local adversarial observations, and solves a sparse, magnitude-bounded FDI synthesis problem over this surrogate via a bi-level projected gradient-based proximal policy optimization (PGPPO) algorithm. The main contributions of our work are summarized as follows.
\begin{itemize}
  \item \textbf{Zero-query, limited information-based adversary model}. We formalize a practical cyber-physical threat model in which the adversary accesses only local topology, regional measurements, and historical outputs of the detector model, without model internals or online query feedback. Moreover, synthesized FDI perturbations are constrained by sparsity and relative magnitude bounds to reflect realistic attack budget constraints and measurement-corruption limits.
  \item \textbf{Heterogeneous distillation with residual calibration}. Because DoS blocks attacked-area measurements at the control center while the adversary observes only a local neighborhood, the surrogate and the detector model operate under asymmetric input views. We combine heterogeneous graph distillation with a second-stage residual calibration module that explicitly corrects this view-induced logit discrepancy, reducing logit-level mean squared error by roughly two-thirds without requiring aligned inputs at deployment.
  \item \textbf{Bi-level PGPPO for sparse and bounded FDI synthesis}. We cast FDI synthesis as a hybrid discrete--continuous problem under a group-balanced logit objective that equally promotes missed detections on attacked lines and false alarms on unattacked candidate lines. PGPPO decouples combinatorial support selection (via an upper-level graph PPO policy) from continuous perturbation refinement (via lower-level projected gradient descent).
\end{itemize}

The remainder of this paper reviews related work in Section~\ref{sec:related}, formulates the system and threat models in Section~\ref{sec:problem}, presents HKD-ACPA and PGPPO in Section~\ref{sec:method}, reports experimental evaluations in Section~\ref{sec:experiments}, and concludes in Section~\ref{sec:conclusion}.

\section{Related Work}
\label{sec:related}

\subsection{Cyber-Physical Attack Detection in Power Grids}
Existing studies have established both model-based and learning-based foundations for power-grid attack detection \cite{hahn2011Cyber,kim2022Survey}. Classical FDI defense methods protect or reconfigure measurements to preserve state-estimation reliability \cite{deng2017Defending,liu2018Reactance,li2022Detection,xu2023Robust}, while PMU-oriented analyses further characterize when FDI attacks can remain unobservable under enhanced bad-data detection \cite{chu2018Unobservable}. Coordinated cyber-physical attack studies show that cyber manipulations can conceal physical line outages and complicate state recovery \cite{soltan2015Joint,deng2017CCPA,soltan2018Power,soltan2019Line,huang2022Link}. Online detection methods have also been developed to distinguish FDI attacks and coordinated cyber-physical attacks from streaming system measurements \cite{zhou2021RealTime}. More recent detector models exploit data-driven and graph-structured representations to localize FDI attacks, line faults, and parallel cyber-physical attacks under partial observations \cite{yan2016Detection,sayghe2020Survey,boyaci2022Joint,fang2023Explainable,fahim2024Graph,wang2025Distribution,ren2026multiscaleattentionbasedattackdiagnosis}. Graph signal processing provides another perspective for modeling spatio-temporal structure in grid measurements, supporting anomaly detection and measurement recovery on power-network graphs \cite{ramakrishna2021GridGraph}. These works mainly aim to improve detection or diagnosis; our focus is complementary, namely how an adversary can synthesize transferable FDI perturbations to confuse and mislead such a detector model.

\subsection{Adversarial Attacks on a Learning-Based Detector Model}
Learning-based power-grid models inherit the adversarial vulnerability of neural networks. Recent studies demonstrate that adversarial perturbations can degrade renewable-energy forecasting and smart-grid state-estimation pipelines \cite{ruan2024On,tian2025EVADE}. In graph learning and cyber-physical systems, black-box and transferable attacks estimate or transfer attack directions without full access to the victim model \cite{bhattacharjee2024DeeBBAA,yu2025GZOO,zhao2024hgattack}, and distillation-based methods show that attack behavior can be compressed into efficient surrogate models \cite{gil2019WhitetoBlack}. 

However, most existing methods assume either white-box model access or repeated online queries. HKD-ACPA instead targets a limited-information setting in which the adversary only has local measurements and historical outputs of the detector model, and uses heterogeneous knowledge distillation plus residual calibration to support sparse, bounded FDI synthesis without online queries to the detector model.

\section{Problem Statement}
\label{sec:problem}
In this section, we introduce the cyber-physical power system model considered in this work and formalize the corresponding adversary model. We first present the physical-layer DC power flow model and then describe the cyber layer that delivers measurements to the control center under potential failures or attacks.

\subsection{Physical Model of Power Grid}
We consider a cyber-physical power system described by the linearized direct-current (DC) power flow model. The power grid is represented by an undirected graph ${\mathcal G} = ({\mathcal V}, {\mathcal E})$, where ${\mathcal V}$ denotes the set of buses and ${\mathcal E}$ denotes the set of transmission lines. For each line $e = (i, j) \in {\mathcal E}$, let $y_{ij} \in \mathbb{R}$ denote its (DC) susceptance and let $\theta_i$ denote the voltage phase angle at bus $i$. The DC line flows and nodal power balance are given by
\begin{align}
  p_{ij} &= y_{ij}(\theta_i - \theta_j), \quad \forall (i,j) \in {\mathcal E}, \label{eq:dc_lineflow}\\
  \sum_{j \in {\mathcal N}_i} p_{ij} &= p_i, \quad \forall i \in {\mathcal V}, \label{eq:dc_balance}
\end{align}
where $p_{ij}$ is the power flow on line $(i,j)$, ${\mathcal N}_i$ is the set of neighbors of bus $i$, and $p_i$ is the net power injection at bus $i$. Under the supply--demand balance condition $\sum_{i \in {\mathcal V}} p_i = 0$ and with a fixed reference bus angle (e.g., $\theta_{r}=0$ for some $r \in {\mathcal V}$), the DC power flow equations can be written compactly as
\begin{equation}
  A \theta = p, \label{eq:power_flow}
\end{equation}
where $A \in \mathbb{R}^{|{\mathcal V}| \times |{\mathcal V}|}$ is the bus susceptance (weighted Laplacian) matrix, $\theta \in \mathbb{R}^{|{\mathcal V}|}$ is the vector of bus voltage angles, and $p \in \mathbb{R}^{|{\mathcal V}|}$ is the vector of power injections. The matrix $A$ admits the factorization $A = D^\top \Gamma D$, where $\Gamma \in \mathbb{R}^{|{\mathcal E}| \times |{\mathcal E}|}$ is a diagonal matrix collecting the line susceptances and $D \in \mathbb{R}^{|{\mathcal E}| \times |{\mathcal V}|}$ is the oriented incidence matrix. The element-wise definitions of $A$ and $D$ are given in Appendix~\ref{app:matrix_def}.

\subsection{Cyber Layer and Adversary Model}
In the cyber layer, field sensors transmit nodal and line measurements to the control center over communication channels that are subject to packet loss and malicious manipulation. To maintain situational awareness under partial or corrupted observations, the control center deploys a learning-based detector model (often an ensemble for improved robustness) to localize attacked lines \cite{ren2026multiscaleattentionbasedattackdiagnosis, huang2022Link, wang2025Distribution}. We consider an intelligent cyber-physical adversary who seeks to induce physical line failures while disrupting this detector model, with the following capabilities:
\begin{itemize}
  \item \textbf{Physical attack capabilities}: The adversary can disconnect transmission lines or alter the susceptances of selected lines within a targeted area of the grid.
  \item \textbf{Cyber attack capabilities}: In the cyber layer, the adversary can block measurements (DoS) or inject malicious data into specific communication channels to corrupt the measurements received by the control center.
  \item \textbf{Knowledge of the power grid}: The adversary has a priori knowledge of the targeted area, including the local topology and line susceptances. Moreover, the adversary can obtain limited historical outputs of the detector model by eavesdropping on communication channels.
\end{itemize}
Within the same monitoring cycle, the adversary compromises a subset of transmission lines in the targeted area and launches coordinated DoS and FDI attacks on transmitted measurements to mislead the detector model. Let ${\mathcal V}_{a} \subseteq {\mathcal V}$ and ${\mathcal E}_{a} \subseteq {\mathcal E}$ denote the targeted buses and candidate transmission lines, following the physical and DoS settings in PCPA \cite{ren2026multiscaleattentionbasedattackdiagnosis} and JCPA \cite{soltan2018Power, soltan2019Line, huang2022Link}. Given only local information around the targeted area, our goal is to synthesize FDI perturbations that degrade physical-attack localization by inducing both missed detections and false alarms.

\subsection{Attack and Detection Problem}
Let $\mathcal{V}_{\bar a}$ and $\mathcal{E}_{\bar a}$ denote the buses and transmission lines free of DoS and physical attacks, respectively. To keep the attack scenario physically and operationally feasible, we impose two standard assumptions.
\begin{assumption}
  \label{asp:attack_size}
  The numbers of attacked buses and transmission lines satisfy
  $|\mathcal{V}_{a}| \leq |\mathcal{V}|/2$ and
  $|\mathcal{E}_{a}| \leq |\mathcal{E}|/2$.
\end{assumption}

\begin{assumption}
  \label{asp:matching}
  There exists a matching in the bipartite subgraph
  $\mathcal{G}[\mathcal{V}_{a}, \mathcal{V}_{\bar a}]$.
\end{assumption}
Assumption~\ref{asp:attack_size} bounds the spatial extent of the attack, while Assumption~\ref{asp:matching} ensures that the control center retains partial observability of the attacked area through boundary measurements and network topology.

Given a set of physically attacked lines $F \subseteq \mathcal{E}_a$, let
\begin{equation}
  \mathbf{X} = [\mathbf{L}^\top, \mathbf{p}^\top, \boldsymbol{\theta}^\top]^\top
\end{equation}
denote the system measurement state collecting bus loads $\mathbf{L} \in \mathbb{R}^{|\mathcal V|}$, power injections $\mathbf{p} \in \mathbb{R}^{|\mathcal V|}$, and voltage phase angles $\boldsymbol{\theta} \in \mathbb{R}^{|\mathcal V|}$. Each candidate line $e \in \mathcal{E}_a$ carries a binary ground-truth label $y_e = \mathbb{I}(e \in F) \in \{0,1\}$, stacked into $\mathbf{y} = [y_e]_{e\in\mathcal{E}_a} \in \{0,1\}^{|\mathcal{E}_a|}$.

At the control center, localization over $\mathcal{E}_a$ is cast as multi-label binary classification trained offline on $(\mathbf{X}_{\bar a}, \mathbf{y})$, where $\mathbf{X}_{\bar a}$ collects measurements not blocked by DoS. Treating the fixed detector model $\mathcal{M}_d$ (whether a single model or an ensemble) as a black-box logit mapping $f_{\mathcal{M}_d}(\cdot)$, its candidate-line logits and detection probabilities are
\begin{equation}
  \mathbf{z}_d = f_{\mathcal{M}_d}(\mathbf{X}_{\bar a}) \in \mathbb{R}^{|\mathcal{E}_a|}, \quad
  \mathbf{P}_a = \sigma(\mathbf{z}_d) \in [0,1]^{|\mathcal{E}_a|},
\end{equation}
where $\sigma(\cdot)$ is the element-wise sigmoid function yielding per-line attack probabilities $P_e = \sigma(z_{d,e})$. Although $\mathcal{M}_d$ may accurately localizes attacked lines from clean $\mathbf{X}_{\bar a}$ \cite{ren2026multiscaleattentionbasedattackdiagnosis, boyaci2022Joint}, the adversary can corrupt accessible channels in $\mathbf{X}_{\bar a}$ via FDI to mislead localization, as developed next.

\section{HKD-ACPA Framework for FDI Synthesis}
\label{sec:method}
\begin{figure*}[ht]
  \vskip 0.05in
  \begin{center}
    \centerline{\includegraphics[width=0.8\textwidth]{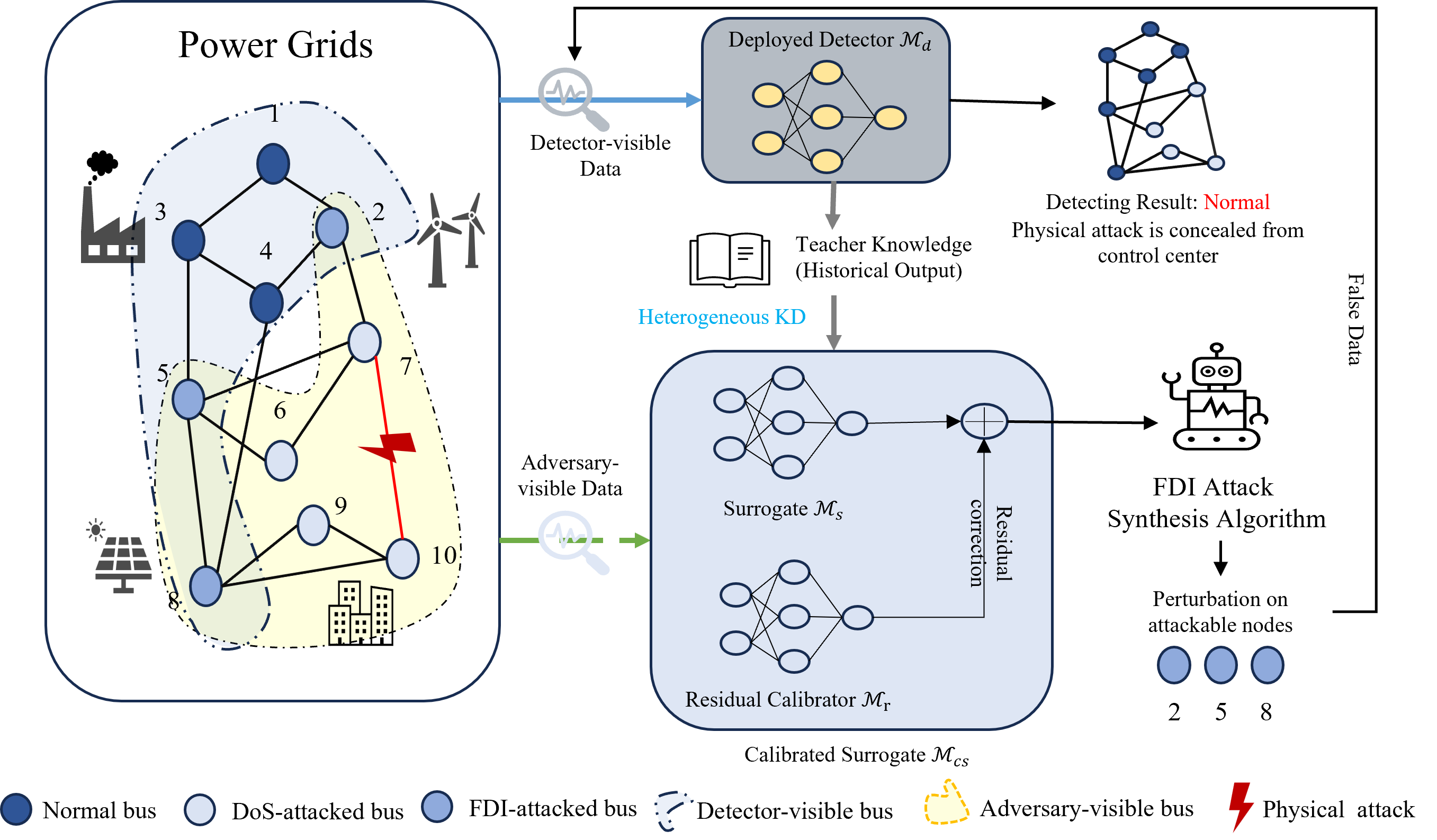}}
    \caption{
      Overview of the proposed heterogeneous knowledge distillation adversarial cyber-physical attack (HKD-ACPA) framework.
    }
    \label{fig: methodoverview}
  \end{center}
  \vskip -0.1in
\end{figure*}

In this section, we present the proposed HKD-ACPA framework, as shown in Fig.~\ref{fig: methodoverview}. The physical attack, DoS attack, and detection problem have been defined above. We focus on constructing a calibrated surrogate model through heterogeneous distillation and residual calibration, and then synthesizing sparse and bounded FDI attacks through a projected gradient-based DRL solver.

\subsection{Distillation Surrogate of the Detector Model}
Prior to launching cyber-physical attacks, we assume that the adversary has access to historical local measurement samples covering the attacked area ($\mathcal{V}_a$) and its neighboring buses ($\mathcal{N}_a$), together with the corresponding candidate-line outputs of the detector model over $\mathcal{E}_a$, obtained from previously observed communication records. These data are used to train a surrogate model $\mathcal{M}_{s}$ that approximates $\mathcal{M}_d$ and provides gradients for generating transferable FDI attacks.

To capture diverse structural patterns, we construct $\mathcal{M}_s$ as an ensemble of $M$ heterogeneous graph-learning base models $\{\mathcal{M}_{s,m}\}_{m=1}^{M}$ with trainable parameters $\Theta_s = \{\Theta_{s,m}\}_{m=1}^{M}$ (reducing to a single model when $M=1$). Due to DoS-induced information asymmetry, $\mathcal{M}_d$ and $\mathcal{M}_s$ operate on different measurement views: $\mathcal{M}_d$ takes $\mathbf{X}_{\bar a}$ as input, whereas each surrogate member $\mathcal{M}_{s,m}$ takes the adversary-accessible local measurements $\mathbf{X}_{\tilde a}$ over $\mathcal{V}_a \cup \mathcal{N}_a$.

Each base surrogate $\mathcal{M}_{s,m}$ is trained by distilling the historical teacher logits $\mathbf{z}_d = f_{\mathcal{M}_d}(\mathbf{X}_{\bar a})$ under the local view $\mathbf{X}_{\tilde a}$. From $\mathbf{z}_d$, we form hard pseudo labels $\hat{\mathbf{y}}_d = \mathbb{I}(\sigma(\mathbf{z}_d) \ge \tau)$ with threshold $\tau$ alongside soft targets $\mathbf{p}=\sigma(\mathbf{z}_d/T)$ at temperature $T$. Letting $\mathbf{z}_{s,m} = f_{\mathcal{M}_{s,m}}(\mathbf{X}_{\tilde a}; \Theta_{s,m}) \in \mathbb{R}^{|\mathcal{E}_a|}$ and $\mathbf{q}=\sigma(\mathbf{z}_{s,m}/T)$, each member $m \in \{1,\dots,M\}$ minimizes a combined hard- and soft-label distillation loss:
\begin{equation}
  \Theta_{s,m}^\star =
  \arg\min_{\Theta_{s,m}}
  \lambda_h
  \mathcal{L}_{\mathrm{BCE}}(\mathbf{z}_{s,m}, \hat{\mathbf{y}}_d)
  +
  \lambda_s T^2
  \mathcal{L}_{\mathrm{KL}}(\mathbf{p}\,\|\,\mathbf{q}),
\end{equation}
where $\lambda_h$ and $\lambda_s$ weight the multi-label binary cross-entropy loss $\mathcal{L}_{\mathrm{BCE}}(\mathbf{z},\tilde{\mathbf{y}}) = -\sum_{e\in\mathcal{E}_a}[\tilde{y}_e \log \sigma(z_e) + (1-\tilde{y}_e)\log(1-\sigma(z_e))]$ and the average Bernoulli KL divergence $\mathcal{L}_{\mathrm{KL}}(\mathbf{p}\,\|\,\mathbf{q}) = \frac{1}{|\mathcal{E}_a|}\sum_{e\in\mathcal{E}_a}[p_e \log(p_e/q_e) + (1-p_e)\log((1-p_e)/(1-q_e))]$. Collecting $\Theta_s^\star = \{\Theta_{s,m}^\star\}_{m=1}^{M}$, the ensemble surrogate $\mathcal{M}_s$ averages the member logits:
\begin{equation}
  \mathbf{z}_s
  = f_{\mathcal{M}_s}(\mathbf{X}_{\tilde a};\Theta_s^\star)
  = \frac{1}{M}\sum_{m=1}^{M} f_{\mathcal{M}_{s,m}}(\mathbf{X}_{\tilde a};\Theta_{s,m}^\star).
\end{equation}

Because $\mathcal{M}_s$ and $\mathcal{M}_d$ operate on asymmetric input views, a systematic logit discrepancy may persist after distillation. We therefore introduce a second-stage residual calibration model $\mathcal{M}_r$ with parameters $\Theta_r$, while freezing both $\mathcal{M}_d$ and $\mathcal{M}_s$. Rather than re-learning $\mathcal{M}_d$ from scratch, $\mathcal{M}_r$ predicts the logit residual $\mathbf{R} = f_{\mathcal{M}_d}(\mathbf{X}_{\bar a}) - f_{\mathcal{M}_s}(\mathbf{X}_{\tilde a};\Theta_s^\star)$ from $\mathbf{X}_{\tilde a}$ via $\hat{\mathbf{R}} = f_{\mathcal{M}_r}(\mathbf{X}_{\tilde a};\Theta_r) \in \mathbb{R}^{|\mathcal{E}_a|}$, yielding the calibrated surrogate logit
\begin{equation}
  \mathbf{z}_{cs}
  = f_{\mathcal{M}_{cs}}(\mathbf{X}_{\tilde a};\Theta_{cs})
  = f_{\mathcal{M}_s}(\mathbf{X}_{\tilde a};\Theta_s^\star)
  + f_{\mathcal{M}_r}(\mathbf{X}_{\tilde a};\Theta_r),
\end{equation}
where $\Theta_{cs} = (\Theta_s^\star, \Theta_r)$. The parameters $\Theta_r$ are learned by minimizing a Smooth $\ell_1$ (Huber) residual loss jointly with a calibrated classification loss:
\begin{equation}
  \Theta_r^\star =
  \arg\min_{\Theta_r}
  \mathcal{L}_{\mathrm{Huber}}(\hat{\mathbf{R}}, \mathbf{R})
  +
  \lambda_b
  \mathcal{L}_{\mathrm{BCE}}(\mathbf{z}_{cs}, \hat{\mathbf{y}}_d),
\end{equation}
where $\lambda_b$ balances logit-residual alignment and pseudo-label consistency, and $\mathcal{L}_{\mathrm{Huber}}(\hat{\mathbf{R}},\mathbf{R}) = \frac{1}{|\mathcal{E}_a|}\sum_{e\in\mathcal{E}_a}\ell_{\mathrm{H}}(\hat{R}_e - R_e)$ uses $\ell_{\mathrm{H}}(u)=\frac{1}{2}u^2$ for $|u|<1$ and $|u|-\frac{1}{2}$ otherwise to limit sensitivity to outlier logit gaps. Setting $\Theta_{cs}^\star = (\Theta_s^\star, \Theta_r^\star)$ defines the calibrated surrogate $\mathcal{M}_{cs}$ used throughout attack synthesis without querying $\mathcal{M}_d$ online.
\begin{remark}
  While the residual calibration losses could in principle be incorporated into the surrogate training objective, we adopt a two-stage formulation to decouple imitation of the detector model from logit-gap correction. The first stage learns a transferable surrogate representation, whereas the second stage freezes the surrogate and learns only the residual discrepancy with respect to the detector model. This residual parameterization provides a more stable and data-efficient calibration mechanism under the distillation process with input-view mismatch.
\end{remark}

\subsection{FDI Attack Synthesis}
To confuse the detector model's localization of physical attacks and mislead the control center, the adversary injects false data into the accessible measurement channels. Let $\mathbf{X}$ denote the original measurement state, and let $\boldsymbol{\delta}$ denote the FDI perturbation applied to the adversary-accessible measurements. The perturbed measurement is
\begin{equation}
  \tilde{\mathbf{X}} = \mathbf{X} + \boldsymbol{\delta},
\end{equation}
where $\boldsymbol{\delta}=[
  \boldsymbol{\delta}_{L}^{\top},
  \boldsymbol{\delta}_{p}^{\top},
  \boldsymbol{\delta}_{\theta}^{\top}
  ]^{\top}$ is constrained by the attackable measurement set and the admissible perturbation budget.

Let $\tilde{\mathbf{X}}_{\tilde a}$ denote the adversary-accessible view of $\tilde{\mathbf{X}}$, with perturbed calibrated logits $\mathbf{z}_{cs}(\tilde{\mathbf{X}}) = f_{\mathcal{M}_{cs}}(\tilde{\mathbf{X}}_{\tilde a};\Theta_{cs}^\star) \in \mathbb{R}^{|\mathcal{E}_a|}$. Let $F\subseteq\mathcal{E}_a$ denote the truly attacked lines and $U=\mathcal{E}_a\setminus F$ the unattacked candidate lines. Defining the groupwise mean calibrated logit $\mu_S(\tilde{\mathbf X}) = |S|^{-1}\sum_{e\in S}z_{cs,e}(\tilde{\mathbf X})$ for $|S|>0$ and $\mu_\emptyset(\tilde{\mathbf X}) = 0$, the attack objective is the difference between the two group means:
\begin{equation}\label{eq:group_objective}
 \mathcal L_{\mathrm{FDI}}(\boldsymbol\delta)
 =\mu_F(\mathbf X+\boldsymbol\delta)-\mu_U(\mathbf X+\boldsymbol\delta),
\end{equation}
so that nonempty groups receive equal group-level weight irrespective of size while an empty group contributes zero. We then formulate the FDI synthesis problem as
\begin{subequations}\label{eq:fdi_attack}
  \begin{align}
    \boldsymbol{\delta}^{\star}
    =&
    \arg\min_{\boldsymbol{\delta}}
    \mathcal L_{\mathrm{FDI}}(\boldsymbol\delta), \\
    \mathrm{s.t.}\quad
    &\mathrm{supp}(\boldsymbol{\delta})
    \subseteq \Omega_a, \quad
    \|\boldsymbol{\delta}\|_0
    \le K, \\
    &|\delta_{i,q}|
    \le
    \epsilon_{i,q},
    \quad \forall (i,q)\in\Omega_a,
  \end{align}
\end{subequations}
where $\Omega_a=\{(i,q)\mid i\in\mathcal{N}_a,\ q\in\mathcal{Q}\}$ denotes the set of attackable measurements over controllable measurement types $\mathcal{Q}$, $K$ is the sparsity budget on the number of modified entries, and $\epsilon_{i,q}>0$ is the admissible magnitude bound for entry $(i,q)$.

Minimizing $\mathcal L_{\mathrm{FDI}}$ in \eqref{eq:group_objective} simultaneously suppresses the calibrated logits on $F$ and raises those on $U$, inducing missed detections and false alarms in parallel while avoiding sigmoid gradient saturation under high-confidence predictions. Normalizing each group by its cardinality further prevents the typically larger unattacked set ($|U| > |F|$) from dominating the optimization direction. Meanwhile, restricting $\Omega_a$ in \eqref{eq:fdi_attack} to the neighboring buses $\mathcal{N}_a$ reflects DoS-induced observability: since measurements on $\mathcal{V}_a$ are blocked from reaching the control center, $\mathcal{N}_a$ forms the local boundary through which adversary-accessible channels influence $\mathcal{M}_d$ within the sparsity budget $K$ and amplitude bounds $\epsilon_{i,q}$.

\subsection{Projected Gradient-based DRL for FDI Attack Synthesis}
The optimization problem in \eqref{eq:fdi_attack} is challenging because the adversary must simultaneously determine \emph{which} measurements to manipulate and \emph{how much} perturbation to inject on the selected entries. The former is a combinatorial support-selection problem under the sparsity budget $\|\boldsymbol{\delta}\|_0 \le K$, whereas the latter is a continuous constrained optimization problem under the box constraints $|\delta_{i,q}| \le \epsilon_{i,q}$. To address this hybrid discrete--continuous structure, we propose a projected gradient-based proximal policy optimization (PGPPO) algorithm.

\begin{figure*}[t]
  \vskip 0.05in
  \begin{center}
    \centerline{\includegraphics[width=0.8\textwidth]{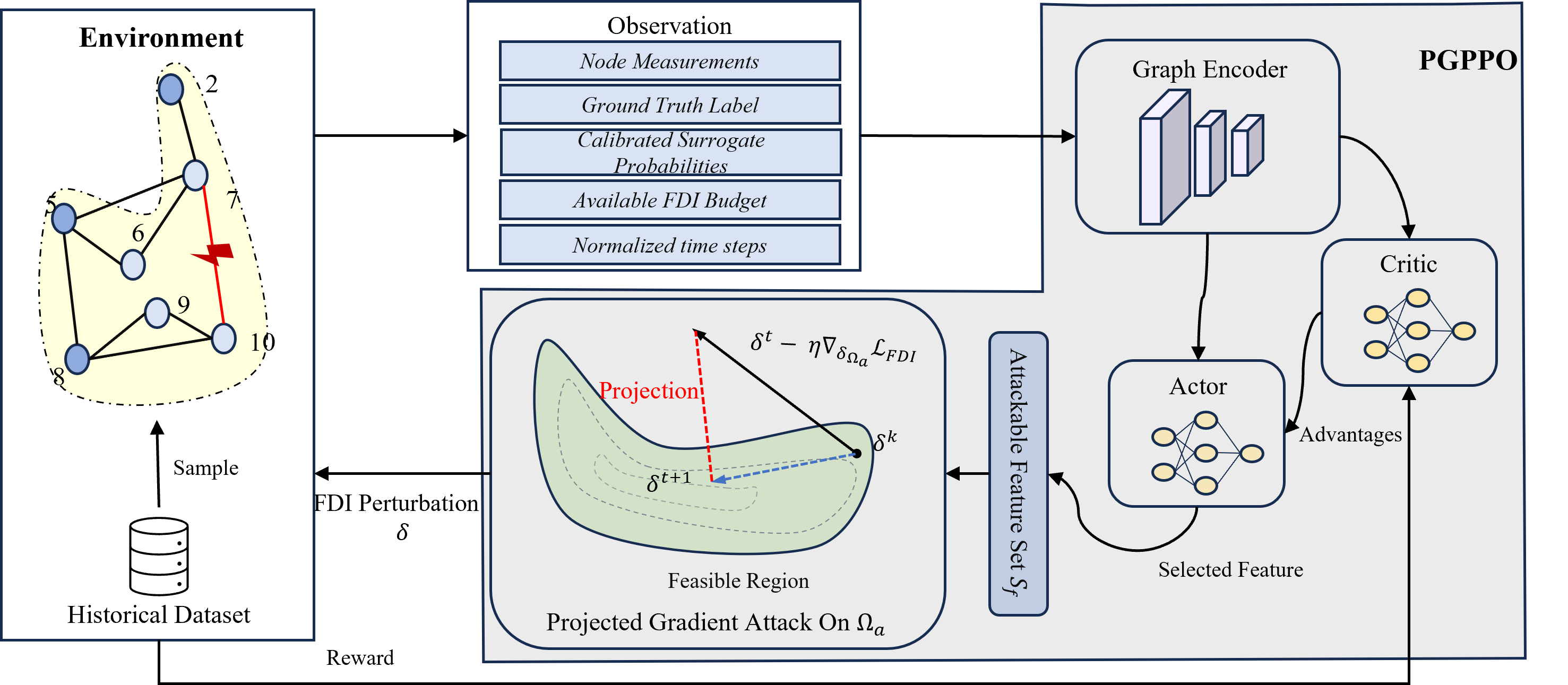}}
    \caption{
      Architecture of the proposed PGPPO algorithm. The upper-level PPO policy learns a combinatorial attack-structure selection strategy, while the lower-level projected-gradient optimizer computes the perturbation magnitudes on the selected support.
    }
    \label{fig: algorithm}
  \end{center}
  \vskip -0.1in
\end{figure*}

The proposed PGPPO adopts a bi-level attack synthesis strategy. At the upper level, a graph encoder-based PPO policy sequentially selects the most influential attackable measurement entries from $\Omega_a$. At the lower level, given the support selected by the policy, a projected gradient procedure optimizes the perturbation magnitudes over the selected entries so as to minimize the group-balanced objective in \eqref{eq:group_objective}. In this way, the RL policy focuses on learning an effective sparse attack structure, while the projected-gradient optimizer solves the downstream continuous perturbation problem efficiently. The overall architecture is shown in Fig.~\ref{fig: algorithm}.

\subsubsection{DRL Formulation}
We cast sparse FDI synthesis as a finite-horizon Markov decision process $(\mathcal{S}, \mathcal{A}, \mathcal{P}, \mathcal{R}, \gamma)$ in which the adversary sequentially selects attackable measurement entries up to horizon $K$. At step $t$, the state summarizes the current attack status:
\begin{equation}
  s_t = \big(\tilde{\mathbf{X}}^t_{\tilde a},\, \mathbf{y},\, \mathbf{p}^t_{cs},\, \boldsymbol{\tau}^t,\, \phi^t \big),
\end{equation}
where $\tilde{\mathbf{X}}^t_{\tilde a}$ denotes the adversary-observable local measurements, $\mathbf{y}$ is the ground-truth line-attack label over $\mathcal{E}_a$, $\mathbf{p}^t_{cs} = \sigma(\mathbf{z}_{cs}(\tilde{\mathbf{X}}^t))$ collects the current calibrated surrogate probabilities, $\boldsymbol{\tau}^t = [|\delta^t_{i,q}| / \epsilon_{i,q}]_{(i, q) \in \Omega_a}$ records the normalized perturbation utilization on $\Omega_a$, and $\phi^t = t/K$ is the normalized step index.

An action $a_t \in \mathcal{A}$ selects one unchosen entry from $\Omega_a$ and updates the binary support mask via $\mathbf{M}_{t+1} = \mathbf{M}_t \vee \mathbf{e}_{a_t}$, where $\mathbf{e}_{a_t}$ is the one-hot indicator of $a_t$. Note that the final number of nonzero entries may be strictly smaller than $K$ when projected updates vanish or early stopping is triggered.

Given $\mathbf{M}_{t+1}$, the lower-level optimizer minimizes $\mathcal{L}_{\mathrm{FDI}}(\boldsymbol{\delta})$ in \eqref{eq:group_objective} over the feasible set $\mathcal{C}(\mathbf{M}_{t+1}) = \{\boldsymbol{\delta} : \mathrm{supp}(\boldsymbol{\delta}) \subseteq \mathrm{supp}(\mathbf{M}_{t+1}),\, |\delta_{i,q}|\le \epsilon_{i,q}\}$ using $\beta$ inner projected-gradient steps. Initializing $\boldsymbol{\delta}^{t,0} = \boldsymbol{\delta}^t$, each inner step $g = 0, \dots, \beta-1$ updates
\begin{equation}\label{eq: projected}
  \boldsymbol{\delta}^{t,g+1}
  =
  \Pi_{\mathcal{C}(\mathbf{M}_{t+1})}
  \Big(
    \boldsymbol{\delta}^{t,g}
    -
    \eta \nabla_{\boldsymbol{\delta}} \mathcal{L}_{\mathrm{FDI}}(\boldsymbol{\delta}^{t,g})
  \Big),
\end{equation}
with step size $\eta$, projection operator $\Pi_{\mathcal{C}(\mathbf{M}_{t+1})}(\cdot)$, and $\boldsymbol{\delta}^{t+1} = \boldsymbol{\delta}^{t,\beta}$. Because both support expansion and inner refinement are deterministic, state transitions are deterministic given $(s_t, a_t)$. The immediate reward measures the step-wise reduction in the calibrated objective,
\begin{equation} \label{eq: reward}
  r_t = \mathcal{L}_{\mathrm{FDI}}(\boldsymbol{\delta}^t) - \mathcal{L}_{\mathrm{FDI}}(\boldsymbol{\delta}^{t+1}),
\end{equation}
rewarding actions that jointly lower attacked-line logits and raise unattacked-line logits. Algorithm~\ref{alg:pgppo} summarizes the training procedure, and Appendix~\ref{app:pgppo} provides the encoder architecture and PPO hyperparameters.
\begin{algorithm}[t]
  \caption{Projected Gradient-based PPO for FDI Attack}
  \label{alg:pgppo}
  \begin{algorithmic}[1]
  \REQUIRE Training samples $\mathcal{D}$, calibrated surrogate model $\mathcal{M}_{cs}$, attackable set $\Omega_a$, candidate line set $\mathcal E_a$, sparsity budget $K$, perturbation bounds $\epsilon$, projected-gradient steps $\beta$, step size $\eta$
  \ENSURE PPO policy $\pi_{\theta}$ for selecting FDI support
  \STATE Initialize actor $\pi_{\theta}$, critic $V_{\phi}$, and trajectory buffer $\mathcal{B}$
  \FOR{each training episode}
      \STATE Sample measurement $\mathbf{X}$ and label $\mathbf{y}$ from $\mathcal{D}$
      \STATE Set $F=\{e\in\mathcal E_a:y_e=1\}$ and $U=\mathcal E_a\setminus F$; omit empty-group terms in Eq.~\eqref{eq:group_objective}
      \STATE Initialize $\boldsymbol{\delta}^{0}\leftarrow \mathbf{0}$ and support mask $\mathbf{M}_{0}\leftarrow \mathbf{0}$
      \FOR{$t=0,\ldots,K-1$}
          \STATE Construct state $s_t=(\tilde{\mathbf{X}}^t_{\tilde a},\mathbf{y},\mathbf{p}^t_{cs},\boldsymbol{\tau}^t,\phi^t)$
          \STATE Select an unchosen measurement entry $a_t\sim\pi_{\theta}(\cdot\mid s_t)$
          \STATE Update support mask $\mathbf{M}_{t+1}\leftarrow \mathbf{M}_{t}\vee\mathbf{e}_{a_t}$ and set $\boldsymbol{\delta}^{t,0}\leftarrow\boldsymbol{\delta}^{t}$
          \FOR{$g=0,\ldots,\beta-1$}
              \STATE Evaluate $\mathcal{L}_{\mathrm{FDI}}(\boldsymbol{\delta}^{t,g})$ via Eq.~\eqref{eq:group_objective} and update $\boldsymbol{\delta}^{t,g+1}$ via Eq.~\eqref{eq: projected}
          \ENDFOR
          \STATE Set $\boldsymbol{\delta}^{t+1}\leftarrow\boldsymbol{\delta}^{t,\beta}$ and $\tilde{\mathbf{X}}^{t+1}\leftarrow\mathbf{X}+\boldsymbol{\delta}^{t+1}$
          \STATE Compute reward $r_t$ via Eq.~\eqref{eq: reward} and store $(s_t,a_t,r_t,s_{t+1})$ in $\mathcal{B}$
      \ENDFOR
      \STATE Update $\pi_{\theta}$ and $V_{\phi}$ using the PPO clipped objective on $\mathcal{B}$, then clear $\mathcal{B}$
  \ENDFOR
  \end{algorithmic}
\end{algorithm}

\section{Experimental Results}
\label{sec:experiments}
In this section, we evaluate HKD-ACPA on cyber-physical attack benchmarks constructed on the IEEE 30-bus and IEEE 118-bus test systems. The experiments are organized around three questions: \textbf{(i)} whether the calibrated surrogate approximates the deployed detector model under asymmetric input views; \textbf{(ii)} whether FDI perturbations synthesized on the surrogate transfer to the deployed detector model; and \textbf{(iii)} how attack effectiveness changes with the feature budget.

\subsection{Dataset and Attack Scenarios}
We generate cyber-physical attack datasets on the IEEE 30-bus ($|\mathcal{V}|=30$, $|\mathcal{E}|=41$) and IEEE 118-bus ($|\mathcal{V}|=118$, $|\mathcal{E}|=186$) systems. Bus loads are randomly sampled from realistic residential load profiles, and power-flow simulations are solved in MATPOWER across the resulting operating points. For each system, the attacked area $(\mathcal{V}_a, \mathcal{E}_a)$ is constructed via a degree-based greedy search satisfying Assumptions~\ref{asp:attack_size} and~\ref{asp:matching} \cite{ren2026multiscaleattentionbasedattackdiagnosis}: $(|\mathcal{V}_a|, |\mathcal{E}_a|) = (8, 8)$ on IEEE 30-bus and $(20, 24)$ on IEEE 118-bus. Within each sample, the physical attack set $F$ is sampled from $\mathcal{E}_a$, where admittance-altering attacks scale nominal line impedances by a random factor in $(1.5, 5)$ and line disconnections apply an impedance factor in $(100, 1000)$ to emulate an open line; simultaneously, DoS attacks block measurements associated with $\mathcal{V}_a$, and HKD-ACPA injects FDI perturbations into the attackable channels $\Omega_a$.

Each sample records the post-attack bus load vector $\mathbf{L}$, power injection vector $\mathbf{p}$, and voltage phase-angle vector $\boldsymbol{\theta}$, paired with the binary label vector $\mathbf{y} \in \{0,1\}^{|\mathcal{E}_a|}$ indicating whether each candidate line is physically attacked. For each system, we generate $60{,}000$ samples and split them into $80\%$ for training, $10\%$ for validation, and $10\%$ for testing.

\subsection{Victim Detector Model, Surrogate, and Baselines}
The deployed detector model $\mathcal{M}_d$ localizes physically attacked transmission lines from the measurements visible to it, $\mathbf{X}_{\bar a}$. To reflect practical control-center deployment, we instantiate the detector model as an ensemble of three learning-based models for cyber-physical attack diagnosis: a graph autoencoder (GAE), a Transformer, and a convolutional graph cross-attention (CGCA) model \cite{fahim2024Graph,thomas2023CNNTransformer,ren2026multiscaleattentionbasedattackdiagnosis}. The adversary does not access the architecture, parameters, or gradients of the detector model, and receives no online query feedback during HKD-ACPA FDI synthesis.

We compare HKD-ACPA with six baselines: Random-FDI, which samples bounded perturbations on random features; Random-PGD, which replaces learned coordinate selection with random selection while retaining the calibrated surrogate and four-step inner optimizer; FGA, a single-step surrogate-gradient attack \cite{chen2018fast}; PGD, an iterative projected-gradient baseline \cite{xu2019topology}; and the online-query methods ZOO \cite{yu2025GZOO} and ALA \cite{ruan2024On}. ZOO and ALA query the deployed detector model during synthesis, whereas HKD-ACPA and the surrogate-based baselines do not.

The surrogate is an ensemble of GAT and GCN components followed by residual calibration; architecture details are given in Appendix~\ref{app:surrogate_arch}. For each case, we sample $1000$ test examples containing at least one physically attacked line without replacement. All methods and all budgets use the same examples within each case. The primary budget is $K=9$, with a sweep over $K\in\{3,6,9,12,15\}$. For every budget, a separate policy is trained for $50{,}000$ episodes and selected by validation performance. We do not reuse the $K=9$ policy at other budgets.

Every method is evaluated under the relative constraint
\begin{equation}\label{eq:relative_budget}
 |\delta_{i,q}|\leq\epsilon_{i,q}
 =0.25\max(|X_{i,q}|,10^{-3}),
\end{equation}
with at most $K$ actually modified features. All perturbations satisfy the measurement-domain constraints and are restricted to the adversary-accessible region. Runtime is measured as the average elapsed time per sample during attack evaluation.

\subsection{Evaluation Metrics}
We evaluate surrogate quality, localization degradation, and budget sensitivity. Surrogate quality is measured by prediction fidelity and logit-level mean squared error (MSE):
\begin{subequations}\label{eq:surrogate_metrics}
\begin{align}
  \mathrm{Fid}
  =&
  \frac{1}{N|\mathcal{E}_a|}
  \sum_{n=1}^{N}
  \sum_{e\in\mathcal{E}_a}
  \mathbb{I}
  \big(
  \hat{y}^{(n)}_{d,e}=\hat{y}^{(n)}_{cs,e}
  \big), \\
  \mathrm{MSE}
  =&
  \frac{1}{N|\mathcal{E}_a|}
  \sum_{n=1}^{N}
  \sum_{e\in\mathcal{E}_a}
  \Big(
  z^{(n)}_{d,e}-z^{(n)}_{cs,e}
  \Big)^2.
\end{align}
\end{subequations}
where $\hat{y}_{d,e}$ and $\hat{y}_{cs,e}$ denote the binary predictions of the deployed detector model and the surrogate model, respectively, and $z_{d,e}$ and $z_{cs,e}$ denote their corresponding logits. Fidelity evaluates agreement after thresholding, whereas MSE measures logit-level alignment, which is more directly related to the gradients used for FDI synthesis.

Attack effectiveness is measured by the decrease in mean sample-level F1, the conditional attack success rate (ASR), and changes in false-negative and false-positive rates (FNR and FPR). Let $\hat y_{d,e}^{(n)}$ and $\hat y_{d,e}^{\prime(n)}$ denote predictions before and after FDI injection. Define $c_n=1$ when all truly attacked lines in sample $n$ are detected before FDI, and $s_n=1$ when at least one such line is missed after FDI. Then
\begin{equation}
 \mathrm{ASR}=\frac{\sum_{n=1}^{N}c_ns_n}{\sum_{n=1}^{N}c_n}.
\end{equation}
Because multi-line attacks require every attacked line in $F$ to be detected prior to FDI for $c_n=1$, the eligible denominators among the $1000$ test samples are $258$ on IEEE 30-bus and $408$ on IEEE 118-bus (a zero denominator yields an undefined ASR).

For sample $n$, let $\mathrm{TP}_n$, $\mathrm{FP}_n$, $\mathrm{FN}_n$, and $\mathrm{TN}_n$ denote the line-level confusion counts. Throughout the paper, reported F1, FNR, and FPR denote sample-averaged metrics computed identically before and after FDI injection:
\begin{align}
 \mathrm{F1}&=\frac1N\sum_{n=1}^N
 \frac{2\mathrm{TP}_n}{2\mathrm{TP}_n+\mathrm{FP}_n+\mathrm{FN}_n},\\
 \mathrm{FNR}&=\frac1N\sum_{n=1}^N
 \frac{\mathrm{FN}_n}{\mathrm{TP}_n+\mathrm{FN}_n},\\
 \mathrm{FPR}&=\frac1{|\mathcal I_U|}\sum_{n\in\mathcal I_U}
 \frac{\mathrm{FP}_n}{\mathrm{FP}_n+\mathrm{TN}_n},
\end{align}
where $\mathcal I_U$ indexes samples with at least one unattacked candidate line so that undefined per-sample FPR values are excluded. We define $\Delta\mathrm{F1}=\mathrm{F1}_{\rm pre}-\mathrm{F1}_{\rm post}$, whereas $\Delta\mathrm{FNR}$ and $\Delta\mathrm{FPR}$ are post-attack minus pre-attack values. F1 captures joint missed-detection and false-alarm degradation, while ASR measures newly induced concealment on initially fully localized attacks. We also report the mean number of modified features and runtime per sample.

\subsection{Main Results}
We evaluate surrogate approximation, transfer attack performance, and budget sensitivity on both IEEE systems; full-test accuracy, precision, and recall of each detector component are given in Appendix~\ref{app:detector_details}.

\subsubsection{Surrogate Approximation}
Table~\ref{tab:case30_surrogate} shows that residual calibration substantially improves logit alignment on both systems, reducing MSE by roughly two-thirds relative to the uncalibrated ensemble. Thresholded fidelity improves on IEEE 118-bus and decreases slightly on IEEE 30-bus, indicating that binary agreement and continuous output alignment capture different properties. For the proposed groupwise logit objective, calibration addresses the continuous discrepancy that remains after distillation, making it a useful complement to the surrogate's imitation of the detector model's decisions.

\begin{table}[t]
\caption{Surrogate quality on the IEEE 30-bus and IEEE 118-bus test sets. Metrics are computed over all candidate lines in $\mathcal{E}_a$.}
\label{tab:case30_surrogate}
\vskip 0.05in
\begin{center}
\begin{small}
\setlength{\tabcolsep}{4pt}
\begin{tabular}{llcc}
\toprule
Case & Model & Fid. $\uparrow$ & MSE $\downarrow$ \\
\midrule
\multirow{4}{*}{IEEE 30-bus} & GAT surrogate & 0.9023 & 8.5169 \\
 & GCN surrogate & 0.8478 & 9.9146 \\
 & Ensemble surrogate $\mathcal{M}_s$  & 0.8981 & 7.9057 \\
 & Calibrated surrogate $\mathcal{M}_{cs}$  & 0.8831 & 2.6813 \\
\midrule
\multirow{4}{*}{IEEE 118-bus} & GAT surrogate & 0.8770 & 5.2328 \\
 & GCN surrogate & 0.7150 & 19.7599 \\
 & Ensemble surrogate $\mathcal{M}_s$  & 0.8470 & 9.5683 \\
 & Calibrated surrogate $\mathcal{M}_{cs}$  & 0.8916 & 2.9326 \\
\bottomrule
\end{tabular}
\end{small}
\end{center}
\vskip -0.1in
\end{table}

\begin{table*}[t]
\caption{Attack results on the $1000$-sample evaluation subsets at $K=9$ and relative bound $0.25$ (including pre-attack No-FDI reference rows).}
\label{tab:main_results}
\centering
\small
\setlength{\tabcolsep}{3.6pt}
\begin{tabular}{llrrrrrr}
\toprule
Case & Method & F1 $\downarrow$ & ASR (\%) $\uparrow$ & FNR & FPR & Features & s/sample \\
\midrule
30 & Pre-attack (No FDI) & 0.7739 & -- & 0.2979 & 0.0521 & 0.000 & -- \\
30 & Random-FDI & 0.7204 & 24.42 & 0.2928 & 0.1769 & 8.428 & 0.033 \\
30 & Random-PGD & 0.7136 & 24.81 & 0.3241 & 0.1438 & 2.932 & 0.806 \\
30 & FGA & 0.5736 & 66.28 & 0.3477 & 0.5254 & 9.000 & 0.058 \\
30 & PGD & 0.6158 & 53.49 & 0.3218 & 0.4271 & 9.000 & 1.921 \\
30 & ZOO & 0.5411 & 70.54 & 0.4256 & 0.4633 & 9.000 & 1.770 \\
30 & ALA & 0.5349 & 68.99 & 0.4510 & 0.4078 & 9.000 & 1.817 \\
30 & \textbf{HKD-ACPA} & \textbf{0.5167} & \textbf{74.03} & 0.4427 & 0.4957 & 6.958 & 1.196 \\
\midrule
118 & Pre-attack (No FDI) & 0.7490 & -- & 0.2529 & 0.3003 & 0.000 & -- \\
118 & Random-FDI & 0.7135 & 25.25 & 0.2608 & 0.3490 & 8.381 & 0.196 \\
118 & Random-PGD & 0.7307 & 6.37 & 0.2612 & 0.3194 & 5.840 & 1.246 \\
118 & FGA & 0.6770 & 6.86 & 0.1228 & 0.6086 & 9.000 & 0.222 \\
118 & PGD & 0.6543 & 44.36 & 0.2177 & 0.5166 & 9.000 & 2.289 \\
118 & ZOO & 0.6556 & 41.42 & 0.3112 & 0.3977 & 9.000 & 15.176 \\
118 & ALA & 0.7031 & 22.06 & 0.2820 & 0.3477 & 9.000 & 8.957 \\
118 & \textbf{HKD-ACPA} & \textbf{0.6267} & \textbf{46.32} & 0.3326 & 0.4204 & 8.846 & 1.256 \\
\bottomrule
\end{tabular}
\end{table*}

\subsubsection{Performance of FDI Attacks}
Table~\ref{tab:main_results} shows that HKD-ACPA achieves the strongest overall localization degradation at the primary budget ($K=9$) on both systems, combining the lowest post-attack F1 with the highest conditional ASR. Its clear lead over FGA and PGD highlights the benefit of coupling support selection with magnitude refinement under sparse budgets, while its large margin over Random-PGD---which shares the exact same calibrated surrogate and four-step inner optimizer---directly validates the learned PPO coordinate selection. Moreover, HKD-ACPA outperforms the online-query baselines ZOO and ALA on both systems without querying $\mathcal{M}_d$ during synthesis; on IEEE 118-bus, it is also an order of magnitude faster per sample than ZOO ($1.256$\,s vs.\ $15.176$\,s). Paired bootstrap tests ($2000$ resamples) on the IEEE 30-bus subset at $K=9$ confirm statistically positive additional F1 decreases of $0.0182$ ($95\%$ CI $[0.0029,0.0335]$) over ALA and $0.0243$ ($95\%$ CI $[0.0100,0.0399]$) over ZOO.

Consistent with the group-balanced objective, HKD-ACPA raises both FNR and FPR relative to the pre-attack baseline, simultaneously concealing attacked lines and inducing false alarms on unattacked lines (with false alarms contributing more prominently on IEEE 30-bus). Notably, HKD-ACPA uses fewer than the $K=9$ permitted features on average ($6.958$ on IEEE 30-bus and $8.846$ on IEEE 118-bus), showing that effective transfer attacks need not exhaust the full support allowance.

\begin{table*}[t]
\caption{F1 decrease ($\Delta\mathrm{F1}$) across feature budgets. Each policy is trained separately for its budget and selected on validation data. Larger values indicate greater degradation of the detector model; bold marks the largest decrease in each row.}
\label{tab:budget_results}
\centering
\small
\setlength{\tabcolsep}{3.6pt}
\begin{tabular}{lrrrrrrrr}
\toprule
Case & $K$ & Random-FDI & Random-PGD & FGA & PGD & ZOO & ALA & HKD-ACPA \\
\midrule
30 & 3 & 0.0267 & 0.0183 & 0.1249 & 0.1183 & 0.1031 & 0.0982 & \textbf{0.2196} \\
30 & 6 & 0.0291 & 0.0366 & 0.1732 & 0.1458 & 0.1780 & 0.1854 & \textbf{0.2402} \\
30 & 9 & 0.0535 & 0.0603 & 0.2003 & 0.1580 & 0.2328 & 0.2389 & \textbf{0.2571} \\
30 & 12 & 0.0614 & 0.0678 & 0.2244 & 0.1802 & \textbf{0.2898} & 0.2865 & 0.2609 \\
30 & 15 & 0.0801 & 0.1011 & 0.2347 & 0.1942 & 0.3075 & \textbf{0.3145} & 0.2611 \\
\midrule
118 & 3 & 0.0123 & 0.0059 & 0.0416 & 0.0410 & 0.0459 & 0.0165 & \textbf{0.0633} \\
118 & 6 & 0.0222 & 0.0126 & 0.0559 & 0.0717 & 0.0709 & 0.0339 & \textbf{0.1011} \\
118 & 9 & 0.0355 & 0.0183 & 0.0720 & 0.0947 & 0.0933 & 0.0459 & \textbf{0.1222} \\
118 & 12 & 0.0483 & 0.0261 & 0.0925 & 0.1093 & 0.1161 & 0.0592 & \textbf{0.1498} \\
118 & 15 & 0.0578 & 0.0309 & 0.0996 & 0.1233 & 0.1375 & 0.0729 & \textbf{0.1691} \\
\bottomrule
\end{tabular}
\end{table*}

\subsubsection{Sensitivity to the Feature Budget}
Table~\ref{tab:budget_results} shows that as $K$ increases over $\{3, 6, 9, 12, 15\}$, HKD-ACPA's F1 decrease grows monotonically from $0.2196$ to $0.2611$ on IEEE 30-bus and from $0.0633$ to $0.1691$ on IEEE 118-bus, accompanied by expanding modified supports from $2.721$ to $9.716$ and from $2.960$ to $13.720$ features, respectively (Appendix~\ref{app:budget_full}, Table~\ref{tab:full_budget}). Across the entire sweep, HKD-ACPA outperforms all surrogate-based baselines at every budget on both systems and surpasses both ZOO and ALA in eight of the ten system--budget settings (all five budgets on IEEE 118-bus and $K \in \{3, 6, 9\}$ on IEEE 30-bus), demonstrating strong zero-query transferability especially under tight sparsity budgets.

The two systems exhibit distinct saturation behaviors as $K$ grows. On the smaller IEEE 30-bus neighborhood ($|\mathcal{E}_a|=8$), HKD-ACPA achieves substantial degradation already at $K=3$ and plateaus beyond $K=9$ (with ASR and FNR stabilizing while FPR rises slightly). Although ZOO and ALA overtake HKD-ACPA at $K \in \{12, 15\}$ by exploiting direct online detector feedback, their marginal $\Delta\mathrm{F1}$ gains also diminish at larger budgets, suggesting an intrinsic scenario saturation threshold that residual surrogate--detector mismatch reaches slightly earlier. Conversely, on the larger IEEE 118-bus subgraph ($|\mathcal{E}_a|=24$), $\Delta\mathrm{F1}$, ASR, FNR, and FPR all continue to rise throughout $K \le 15$.

Support utilization corroborates this contrast: at $K=15$, the IEEE 30-bus policy modifies only $9.716$ features on average because extra perturbations yield negligible surrogate gain, whereas the IEEE 118-bus policy utilizes $13.720$ features. Practically, attack budget allocation should therefore account for neighborhood scale and saturation rather than indiscriminately maximizing modified channels.

These findings also suggest practical countermeasures against HKD-ACPA. Encrypting or coarsening diagnostic confidence outputs can impede surrogate distillation and residual calibration, while authenticating high-leverage boundary buses in $\mathcal{N}_a$ and coupling GNN localization with power-flow consistency checks and adversarial boundary training can shrink $\Omega_a$ and expose physically inconsistent injections.

\section{Conclusion}
\label{sec:conclusion}
This paper proposed HKD-ACPA to disrupt learning-based transmission-line attack localization in cyber-physical power grids without model internals or online queries. To bridge the input-view asymmetry between DoS-masked control-center telemetry and local adversarial observations, HKD-ACPA combines heterogeneous graph distillation with second-stage residual logit calibration, and solves sparse, amplitude-bounded FDI synthesis via bi-level PGPPO under a group-balanced logit objective.

Our results reveal that coordinated DoS and sparse boundary FDI inflict compounded, bidirectional diagnostic damage, simultaneously concealing genuine line outages and fabricating false alarms on intact lines to delay fault isolation and misdirect operator dispatch. Because such disruption stems from perturbing only a few channels around the DoS-masked zone, partial-observability boundaries act as critical vulnerability bottlenecks whose marginal damage saturates once dominant channels are exploited. Future work will focus on the fundamental detectability conditions and physics-informed detection of such coordinated hybrid cyber-physical attacks.

\bibliography{IEEE_abb, main}
\bibliographystyle{icml2026}

\clearpage
\appendix
\section{DC Power Flow Matrix Definitions}
\label{app:matrix_def}
In \eqref{eq:power_flow}, the bus susceptance matrix $A \in \mathbb{R}^{|\mathcal{V}|\times|\mathcal{V}|}$ has off-diagonal entries $A_{ij}=-y_{ij}$ for $(i,j)\in\mathcal{E}$ ($A_{ij}=0$ otherwise) and diagonal entries $A_{ii}=\sum_{k\in\mathcal{N}_i} y_{ik}$. In $A = D^\top \Gamma D$, $\Gamma = \mathrm{diag}(y_e)_{e\in\mathcal{E}}$, and the oriented incidence matrix $D \in \mathbb{R}^{|\mathcal{E}|\times|\mathcal{V}|}$ has entries $D_{ei}=1$ if bus $i$ is the sending end of line $e$, $D_{ei}=-1$ if bus $i$ is the receiving end, and $D_{ei}=0$ otherwise.

\section{Model Architectures and Hyperparameters}
\subsection{Surrogate Ensemble and Residual Calibrator}
\label{app:surrogate_arch}
The ensemble $\mathcal{M}_s$ averages logits from $M=2$ members taking three node features (voltage angle, load, power injection): a $3$-layer GCN ($\mathrm{GCNConv}$ widths $(128,128,128)$) and a $2$-layer GAT ($\mathrm{GATv2Conv}$ channels $(64,64)$, heads $(2,2)$, concatenation disabled), both using LayerNorm, LeakyReLU, dropout $p=0.2$, and an endpoint-concatenation MLP head ($2d\to256\to128\to1$) over $\mathcal{E}_a$. Members are trained with AdamW ($\text{lr}=10^{-3}$, weight decay $10^{-4}$, batch size $128$), StepLR (step $50$, $\gamma=0.25$), $(\lambda_h,\lambda_s)=(1.0,0.5)$, $T=1.0$, and $\tau=0.5$. The residual calibrator $\mathcal{M}_r$ uses two $\mathrm{GCNConv}$ layers of widths $(128,128)$ with LayerNorm, LeakyReLU, dropout $p=0.1$, and an MLP head ($256\to256\to128\to1$), trained for $100$ epochs (with $\mathcal{M}_d,\mathcal{M}_s$ frozen) via AdamW ($\text{lr}=10^{-3}$, weight decay $10^{-4}$, batch size $128$, gradient clip $1.0$) and $\lambda_b=0.2$.

\subsection{PGPPO Network and Optimization}
\label{app:pgppo}
Actor and critic use separate encoders combining a node branch (linear layer to $128$ plus two $\mathrm{GCNConv}(128,128)$ layers) and a global branch (two-layer MLP of width $128$), fused via conditional attention pooling (mean, max, attention) into a $384$-dimensional vector. The actor ($384\to256\to64\to|\Omega_a|$) outputs a masked categorical policy, and the critic ($384\to512\to256\to1$) estimates state values; all layers use ReLU, LayerNorm, and Xavier-uniform initialization. Each budget-specific policy is trained for $50{,}000$ episodes (seed $42$, best validation checkpoint retained), updating every $200$ episodes on a $4000$-transition buffer ($5$ epochs, minibatch $512$, $\gamma=0.99$, GAE $\lambda=0.95$, clip $0.2$, entropy $0.01$, AdamW $\text{lr}=10^{-3}$, weight decay $0.01$, gradient clip $0.5$), with inner PGD using $\beta=4$ and $\eta=10^{-3}$.

\section{Detector Performance and Full Budget Sweep}
\label{app:detector_details}
\label{app:budget_full}
Table~\ref{tab:appendix_detector_details} reports full-test pre-attack metrics of $\mathcal{M}_d$ and its base models (evaluation-subset baselines are in Table~\ref{tab:main_results}). Table~\ref{tab:full_budget} provides the complete budget sweep across $1000$ test samples per case (ASR denominators: $258$ on 30-bus, $408$ on 118-bus).

\begin{table}[h!]
\caption{Pre-attack detector metrics on the full test sets.}
\label{tab:appendix_detector_details}
\vskip 0.03in
\centering
\footnotesize
\setlength{\tabcolsep}{2.2pt}
\begin{tabular}{lccc@{\hspace{6pt}}ccc}
\toprule
 & \multicolumn{3}{c}{IEEE 30-bus} & \multicolumn{3}{c}{IEEE 118-bus} \\
\cmidrule(lr){2-4}\cmidrule(lr){5-7}
Model & Acc. & Prec. & Rec. & Acc. & Prec. & Rec. \\
\midrule
CGCA & 0.8290 & 0.9434 & 0.7002 & 0.8655 & 0.8535 & 0.8835 \\
GAE & 0.7739 & 0.8444 & 0.6717 & 0.8185 & 0.8145 & 0.8262 \\
Transformer & 0.7890 & 0.9046 & 0.6462 & 0.8318 & 0.8224 & 0.8477 \\
Ensemble & 0.8216 & 0.9337 & 0.6924 & 0.8534 & 0.8422 & 0.8709 \\
\bottomrule
\end{tabular}
\vskip -0.1in
\end{table}

\begin{table*}[t]
\caption{Complete budget evaluation on the IEEE 30-bus and IEEE 118-bus systems ($1000$ samples per row; $\Delta\mathrm{FNR}$ and $\Delta\mathrm{FPR}$ denote post-attack minus pre-attack values; runtime in s/sample is execution-specific).}
\label{tab:full_budget}
\centering
\scriptsize
\setlength{\tabcolsep}{2.1pt}
\begin{tabular}{rlrrrrrrr@{\hspace{7pt}}rrrrrrr}
\toprule
 & & \multicolumn{7}{c}{IEEE 30-bus} & \multicolumn{7}{c}{IEEE 118-bus} \\
\cmidrule(lr){3-9}\cmidrule(lr){10-16}
$K$ & Method & Post-F1 & FNR & $\Delta$FNR & FPR & $\Delta$FPR & Feat. & Time & Post-F1 & FNR & $\Delta$FNR & FPR & $\Delta$FPR & Feat. & Time \\
\midrule
3 & Random-FDI & 0.7472 & 0.3020 & 0.0040 & 0.1057 & 0.0537 & 2.802 & 0.035 & 0.7367 & 0.2580 & 0.0051 & 0.3138 & 0.0135 & 2.794 & 0.208 \\
3 & Random-PGD & 0.7556 & 0.3065 & 0.0085 & 0.0780 & 0.0259 & 1.024 & 0.322 & 0.7431 & 0.2576 & 0.0048 & 0.3047 & 0.0044 & 1.975 & 0.532 \\
3 & FGA & 0.6489 & 0.3034 & 0.0055 & 0.3617 & 0.3097 & 3.000 & 0.055 & 0.7074 & 0.1799 & -0.0730 & 0.4715 & 0.1712 & 3.000 & 0.222 \\
3 & PGD & 0.6556 & 0.3207 & 0.0227 & 0.3111 & 0.2590 & 3.000 & 1.907 & 0.7080 & 0.2365 & -0.0164 & 0.3927 & 0.0924 & 3.000 & 0.970 \\
3 & ZOO & 0.6708 & 0.3338 & 0.0359 & 0.2493 & 0.1973 & 3.000 & 1.764 & 0.7030 & 0.2825 & 0.0297 & 0.3453 & 0.0450 & 3.000 & 15.176 \\
3 & ALA & 0.6757 & 0.3514 & 0.0534 & 0.1905 & 0.1384 & 3.000 & 1.810 & 0.7325 & 0.2640 & 0.0111 & 0.3225 & 0.0222 & 3.000 & 8.945 \\
3 & HKD-ACPA & 0.5543 & 0.4409 & 0.1429 & 0.3520 & 0.2999 & 2.721 & 0.381 & 0.6857 & 0.2911 & 0.0383 & 0.3467 & 0.0464 & 2.960 & 0.520 \\
\midrule
6 & Random-FDI & 0.7448 & 0.2830 & -0.0149 & 0.1459 & 0.0938 & 5.597 & 0.034 & 0.7268 & 0.2564 & 0.0035 & 0.3306 & 0.0303 & 5.590 & 0.162 \\
6 & Random-PGD & 0.7373 & 0.3114 & 0.0134 & 0.1150 & 0.0630 & 1.990 & 0.691 & 0.7364 & 0.2583 & 0.0055 & 0.3133 & 0.0130 & 3.918 & 0.920 \\
6 & FGA & 0.6007 & 0.3238 & 0.0259 & 0.4774 & 0.4253 & 6.000 & 0.058 & 0.6931 & 0.1388 & -0.1141 & 0.5497 & 0.2494 & 6.000 & 0.221 \\
6 & PGD & 0.6281 & 0.3160 & 0.0181 & 0.3947 & 0.3427 & 6.000 & 1.933 & 0.6773 & 0.2241 & -0.0288 & 0.4599 & 0.1596 & 6.000 & 1.002 \\
6 & ZOO & 0.5958 & 0.3862 & 0.0882 & 0.3684 & 0.3164 & 6.000 & 1.732 & 0.6781 & 0.2991 & 0.0462 & 0.3694 & 0.0691 & 6.000 & 10.360 \\
6 & ALA & 0.5885 & 0.4169 & 0.1189 & 0.3065 & 0.2545 & 6.000 & 1.773 & 0.7150 & 0.2763 & 0.0235 & 0.3353 & 0.0350 & 6.000 & 8.775 \\
6 & HKD-ACPA & 0.5337 & 0.4401 & 0.1421 & 0.4301 & 0.3780 & 5.274 & 0.721 & 0.6479 & 0.3156 & 0.0627 & 0.3981 & 0.0978 & 5.925 & 0.897 \\
\midrule
9 & Random-FDI & 0.7204 & 0.2928 & -0.0051 & 0.1769 & 0.1248 & 8.428 & 0.033 & 0.7135 & 0.2608 & 0.0079 & 0.3490 & 0.0487 & 8.381 & 0.196 \\
9 & Random-PGD & 0.7136 & 0.3241 & 0.0262 & 0.1438 & 0.0917 & 2.932 & 0.806 & 0.7307 & 0.2612 & 0.0084 & 0.3194 & 0.0191 & 5.840 & 1.246 \\
9 & FGA & 0.5736 & 0.3477 & 0.0497 & 0.5254 & 0.4733 & 9.000 & 0.058 & 0.6770 & 0.1228 & -0.1301 & 0.6086 & 0.3083 & 9.000 & 0.222 \\
9 & PGD & 0.6158 & 0.3218 & 0.0239 & 0.4271 & 0.3750 & 9.000 & 1.921 & 0.6543 & 0.2177 & -0.0351 & 0.5166 & 0.2163 & 9.000 & 2.289 \\
9 & ZOO & 0.5411 & 0.4256 & 0.1277 & 0.4633 & 0.4112 & 9.000 & 1.770 & 0.6556 & 0.3112 & 0.0583 & 0.3977 & 0.0974 & 9.000 & 15.176 \\
9 & ALA & 0.5349 & 0.4510 & 0.1530 & 0.4078 & 0.3558 & 9.000 & 1.817 & 0.7031 & 0.2820 & 0.0292 & 0.3477 & 0.0474 & 9.000 & 8.957 \\
9 & HKD-ACPA & 0.5167 & 0.4427 & 0.1448 & 0.4957 & 0.4436 & 6.958 & 1.196 & 0.6267 & 0.3326 & 0.0798 & 0.4204 & 0.1201 & 8.846 & 1.256 \\
\midrule
12 & Random-FDI & 0.7124 & 0.2883 & -0.0096 & 0.2107 & 0.1586 & 11.205 & 0.034 & 0.7007 & 0.2651 & 0.0123 & 0.3616 & 0.0613 & 11.153 & 0.207 \\
12 & Random-PGD & 0.7061 & 0.3183 & 0.0204 & 0.1765 & 0.1244 & 4.021 & 1.223 & 0.7229 & 0.2667 & 0.0138 & 0.3260 & 0.0257 & 7.710 & 1.614 \\
12 & FGA & 0.5495 & 0.3719 & 0.0739 & 0.5616 & 0.5095 & 11.974 & 0.056 & 0.6565 & 0.1278 & -0.1250 & 0.6529 & 0.3526 & 12.000 & 0.219 \\
12 & PGD & 0.5937 & 0.3411 & 0.0431 & 0.4640 & 0.4119 & 12.000 & 1.897 & 0.6397 & 0.2182 & -0.0347 & 0.5543 & 0.2540 & 12.000 & 1.849 \\
12 & ZOO & 0.4841 & 0.4807 & 0.1828 & 0.5452 & 0.4932 & 12.000 & 1.764 & 0.6329 & 0.3256 & 0.0727 & 0.4276 & 0.1273 & 12.000 & 11.024 \\
12 & ALA & 0.4874 & 0.4947 & 0.1968 & 0.4833 & 0.4312 & 12.000 & 1.742 & 0.6898 & 0.2922 & 0.0394 & 0.3613 & 0.0610 & 12.000 & 8.583 \\
12 & HKD-ACPA & 0.5129 & 0.4443 & 0.1464 & 0.5071 & 0.4550 & 8.174 & 1.163 & 0.5992 & 0.3499 & 0.0971 & 0.4584 & 0.1581 & 11.672 & 1.643 \\
\midrule
15 & Random-FDI & 0.6938 & 0.2949 & -0.0030 & 0.2417 & 0.1896 & 14.022 & 0.033 & 0.6912 & 0.2630 & 0.0102 & 0.3807 & 0.0804 & 13.949 & 0.196 \\
15 & Random-PGD & 0.6728 & 0.3402 & 0.0423 & 0.2169 & 0.1648 & 4.785 & 1.391 & 0.7181 & 0.2679 & 0.0150 & 0.3330 & 0.0327 & 9.686 & 2.010 \\
15 & FGA & 0.5392 & 0.3839 & 0.0859 & 0.5738 & 0.5218 & 14.621 & 0.060 & 0.6494 & 0.1154 & -0.1375 & 0.6896 & 0.3892 & 15.000 & 0.223 \\
15 & PGD & 0.5797 & 0.3501 & 0.0522 & 0.4911 & 0.4390 & 15.000 & 1.885 & 0.6257 & 0.2148 & -0.0381 & 0.5963 & 0.2960 & 15.000 & 2.528 \\
15 & ZOO & 0.4664 & 0.4918 & 0.1938 & 0.5933 & 0.5412 & 15.000 & 1.810 & 0.6115 & 0.3407 & 0.0878 & 0.4543 & 0.1540 & 15.000 & 11.651 \\
15 & ALA & 0.4594 & 0.5170 & 0.2191 & 0.5283 & 0.4762 & 15.000 & 1.762 & 0.6761 & 0.3039 & 0.0510 & 0.3762 & 0.0759 & 15.000 & 8.593 \\
15 & HKD-ACPA & 0.5127 & 0.4425 & 0.1445 & 0.5153 & 0.4633 & 9.716 & 1.544 & 0.5799 & 0.3679 & 0.1151 & 0.4734 & 0.1731 & 13.720 & 1.989 \\
\bottomrule
\end{tabular}
\end{table*}

\end{document}